\documentclass[10pt,letterpaper]{article}

\usepackage{cogsci}

\cogscifinalcopy 

\usepackage{pslatex}
\usepackage{apacite}
\usepackage{booktabs}
\usepackage{bm}
\usepackage{float} 

\usepackage{graphicx}
\usepackage{amsmath,amssymb,amsthm}
\usepackage{enumerate}

\newcounter{list-counter}
\newcommand{\key}{\textbf}

\title{A hierarchical Bayesian model for syntactic priming}
 
\author{{\large \bf Weijie Xu (weijie.xu@uci.edu)} \\
  Department of Language Science, 3151 Social Science Plaza \\
  Irvine, CA 92697 USA
  \AND {\large \bf Richard Futrell (rfutrell@uci.edu)} \\
  Department of Language Science, 3151 Social Science Plaza \\
  Irvine, CA 92697 USA}

\begin{document}

\maketitle

\begin{abstract}
The effect of syntactic priming exhibits three well-documented empirical properties: the lexical boost, the inverse frequency effect, and the asymmetrical decay. We aim to show how these three empirical phenomena can be reconciled in a general learning framework, the hierarchical Bayesian model (HBM). The model represents syntactic knowledge in a hierarchical structure of syntactic statistics, where a lower level represents the verb-specific biases of syntactic decisions, and a higher level represents the abstract bias as an aggregation of verb-specific biases. This knowledge is updated in response to experience by Bayesian inference. In simulations, we show that the HBM captures the above-mentioned properties of syntactic priming. The results indicate that some properties of priming which are usually explained by a residual activation account can also be explained by an implicit learning account. We also discuss the model's implications for the lexical basis of syntactic priming.


\end{abstract}

\section{Introduction}

Syntactic priming refers to an adaptive behavior where exposure to a specific syntactic structure influences a language user's subsequent processing of similar structures. From the production side, speakers tend to repeat the syntactic structure they have recently been exposed to \cite{bock1986syntactic, pickering2008structural, bock1990framing, branigan2000syntactic}. For example, after a person produces or perceives a sentence such as ``the boy gave the girl the ball'' (called the \key{prime}), there is a reliable increase in the probability that their description of a subsequent ditransitive event (called the \key{target}) will use the same syntactic form, in this case using the double-object construction (DO; as in ``give $X$ $Y$'') rather than a prepositional-object construction (PO; as in ``give $X$ to $Y$'').


\subsection{Empirical Background}

A large body of work has examined factors that modulate the strength of syntactic priming \cite{mahowald2016meta}. In (\ref{empirical-phenomena}) we summarize three major findings from this literature, and review them below. 

\begin{enumerate}[(1)]
\setcounter{enumi}{\value{list-counter}}
    \item \label{empirical-phenomena} Three empirical properties of syntactic priming
    \begin{enumerate}[a.]
        \item \textit{Lexical Boost}: Lexical alignment, especially verb overlap, between the prime and the target enhances the priming effect size.
        \item \textit{Inverse Frequency}: Less frequent prime constructions elicit stronger priming effect.
        \item \textit{Asymmetrical Decay}: Lexical boost decays faster than the abstract verb-independent priming effect.
    \end{enumerate}
\setcounter{list-counter}{\value{enumi}}
\end{enumerate}

The \key{lexical boost} refers to an increase in the effect size of syntactic priming when the prime sentence and the target sentence share critical lexical items \cite{pickering1998representation, gries2005syntactic}. For example, if both the prime and the target use the same verb ``give'', then priming of the DO construction is stronger than if the prime and target involved different verbs. In comprehension, the alignment of verb lexical items has been found to be necessary to observe a reliable priming effect \cite{tooley2010syntactic}.

Another well-documented property of syntactic priming is the \key{inverse frequency effect}, where constructions that are less frequent in language experience are susceptible to a stronger priming effect \cite{scheepers2003syntactic, kaschak2011structural, hartsuiker1998syntactic, bock1986syntactic, ferreira2003persistence}. For example, in English usage, PO constructions are overall rarer than DO; consequently, a PO prime causes a larger increase in the probability of a PO target than a DO prime does for a DO target. A more precise characterization of the inverse frequency effect has been put forward by \citeA{jaeger2013alignment}, who claim that the effect size of priming is directly proportional to the prediction error during processing, which in turn depends on statistical regularities in the language experience. In that case, infrequent constructions or co-occurrences cause larger prediction errors and thus stronger priming effect.

Our last phenomenon of interest has to do with the decay of the size of the priming effect as a function of time between the prime and target. Decay of syntactic priming has been observed in both the short-term \cite{levelt1982surface, branigan1999syntactic} and the long-term \cite{bock2007persistent, bock2000persistence, kaschak2011long}, in a way that depends on lexical alignment between the prime and target. In particular, \key{asymmetrical decay} refers to the fact that the lexical boost decays more rapidly than the abstract priming effect, where there is no lexical alignment \cite{bock2000persistence, hartsuiker2008syntactic}. 

\subsection{Related Work}

Two lines of theories are proposed in the literature, the residual activation account and the implicit learning account. Each account captures some but not all the empirical properties mentioned above. 

The \key{residual activation} account of priming holds that production or comprehension of a syntactic form is influenced by its level of \emph{activation}. Upon encountering a sentence, activation is boosted for the representation of both the lexical items and the syntactic structure, facilitating future reuse \cite{pickering1998representation}. Based on a spreading activation model of lemma retrieval \cite{roelofs1992spreading}, \citeA{pickering1998representation} specify a representational architecture where nodes representing verb lemmas are connected to the combinatorial nodes representing syntactic structures. In this model, a boost in activation on a certain verb lemma spreads outward to the combinatorial nodes and to other verb lemmas, which in turn further contribute to the activation of the combinatorial nodes. This residual activation account predicts the the lexical boost effect, since the the verb lemma used in the prime is directly activated and receives the strongest activation boost. However, the residual activation account has difficulty explaining long-term priming, since all activation is posited to decay over time.

The \key{implicit learning} account proposes that syntactic priming is a kind of syntactic learning. This idea was originally implemented in a connectionist model that predicts the next word based on the previously heard word \cite{chang2006becoming}. In this model, prediction error serves as an explicit learning signal, based on which the weights between nodes are updated via backpropagation. The implicit learning account predicts that the priming effect size is proportional to the prediction error when processing the prime, thus accounting for the inverse frequency effect \cite{jaeger2013alignment}. This error-based effect is also reproduced in learning models that do not include an explicit error signal, such as the Bayesian belief-update \cite{kleinschmidt2012belief, fine2013rapid,gershman2015unifying}. However, compared to the residual activation account, the implicit learning account struggles to capture the observation that the syntactic priming effect can sometimes be short-lived. 

Since neither account simultaneously explains all the three properties in (\ref{empirical-phenomena}), some studies propose a hybrid account, holding that both residual activation and implicit learning play a role. This amalgamation is also motivated by the observation of asymmetrical decay. In particular, residual activation explains the rapid decay of lexical boost, whereas long-lived abstract priming is mostly driven by implicit learning \cite{hartsuiker2008syntactic, bock2000persistence}. In line with this hybrid mechanism, \citeA{reitter2011computational} propose an activation-based model in the ACT-R framework \cite{lewis2005activation}, where lexical chunks and syntactic chunks are represented separately in the declarative memory. The priming effect results both from an increase of the base-level activation of the syntactic chunk and from the spreading activation from lexical chunks. Similar to \citeA{pickering1998representation}, the spreading activation mechanism allows the model to capture lexical boost. The model also captures the inverse frequency effect because of the diminished base-level learning for high-frequency chunks. Moreover, the model captures the asymmetrical decay since the spreading activation in ACT-R only goes from chunks that are temporarily held in the working memory buffer to chunks in long-term memory. Therefore, when lexical chunks are evacuated from working memory, they cannot spread activation anymore.

\subsection{The Current Study}

We develop a computational-level theory of syntactic priming in terms of a hierarchical Bayesian model (HBM) of syntactic knowledge. Our HBM extends the basic Bayesian belief-update model by representing syntactic knowledge at multiple levels of abstraction \cite{kemp2007learning}. As a general learning framework, HBMs have been broadly applied to speech perception \cite{kleinschmidt2015robust}, the emergence of communicative systems \cite{hawkins2023partners}, and human cognition in general \cite{tenenbaum2011grow}. We show how this general learning framework can capture the three empirical properties of priming in (\ref{empirical-phenomena}), namely the lexical boost effect, the inverse frequency effect, and asymmetrical decay.

\section{Modeling Framework}

\subsection{Representing Syntactic Knowledge}

The core of the priming effect is a decision-making problem: the decision between multiple syntactic frames for the same message in production, or between multiple parses for an ambiguous input in comprehension. The decision-making can be characterized by a probability distribution over the syntactic alternatives of interest. In the production of ditransitives, for instance, the decision between dative object (DO) and prepositional object (PO) can be captured by a Bernoulli distribution, whose parameter represents the production probability for DO versus PO. We aim to characterize this probability distribution and how it changes in response to experience.

\begin{figure}
    \centering
    \includegraphics[width=0.9\linewidth]{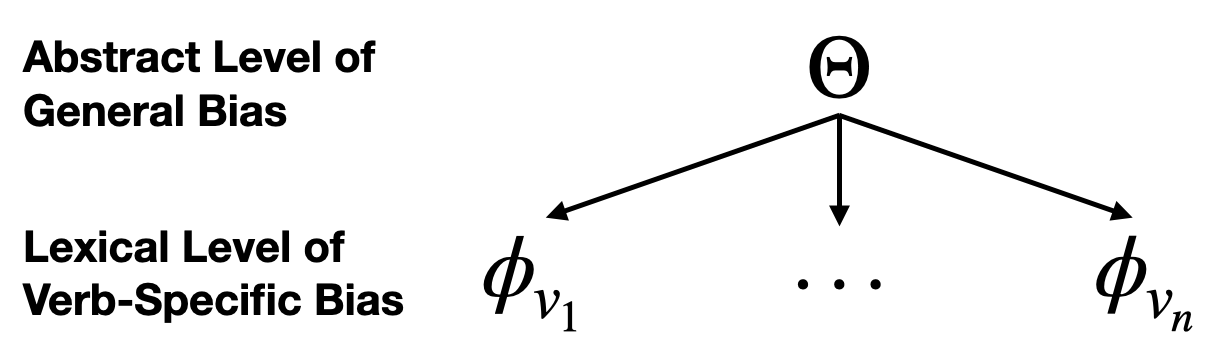}
    \caption{Hierarchical representation of syntactic statistics.} 
    \label{fig:HBM-representation}
\end{figure}

Following previous work, we represent a person's syntactic knowledge as syntactic statistics: counts of how often one expects to experience different syntactic constructions
\cite{fine2013rapid, jaeger2013alignment}, represented at different levels of abstraction. We posit a hierarchical structure for syntactic statistics, as illustrated in Figure~\ref{fig:HBM-representation}. We specify two levels of abstraction. The lower level captures verb-specific statistics $\phi_v$, such that each verb imposes its own bias \cite{bernolet2010does}. The verb-specific statistics echo the lexicalist view of sentence processing, where syntactic knowledge is accessible via lexical representations \cite{macdonald1994lexical, mcrae1998modeling, spivey1995resolving}. 
The higher-level abstraction $\Theta$ is an aggregation over all the lower-level statistics and represents the general decision bias across all the verbs.

Taking ditransitive priming as an example, the variables in Figure~\ref{fig:HBM-representation} would represent counts over DO constructions and PO constructions, conditional on specific verbs (the $\phi_v$) or aggregating across verbs (the high-level $\Theta$). However, this hierarchical structure of syntactic statistics is not limited to any specific construction (e.g. ditransitives). The idea can be applied to any scenario that includes processing decisions, such as ambiguity resolution (for example, garden-path sentences) in comprehension and the choice of syntactic frames in production (for example, passive vs. active for transitives).


\subsection{Updating Syntactic Knowledge}

We posit that syntactic learning takes the form of Bayesian updates to the hierarchical model in Figure~\ref{fig:HBM-representation}. We explain how this works first using a familiar non-hierarchical Beta-Binomial model, and then the full hierarchical model.

\subsubsection{Beta-Binomial Process.} The Beta-Binomial model is a simple version of Bayesian belief updating. It postulates a generative model where a set of decisions $x$ between two outcomes are sampled from a Binomial distribution parameterized by variable $\Theta$, which in turn is sampled from a prior Beta distribution with hyperparameters $\alpha, \beta > 0$:
\begin{align}
    \Theta &\sim \mathrm{Beta}(\alpha, \beta) \label{HBM1-prior} \\
    x &\sim \mathrm{Binomial}(\Theta). \label{HBM1-likelihood}
\end{align}
Applied to syntactic priming, the variable $\Theta$ represents the probability of the DO form versus the PO form, and the hyperparameters $\alpha, \beta$ in the Beta distribution represent prior expectations for frequencies of the DO and PO outcome respectively. After observing an outcome $x$, the variable $\Theta$ updates by Bayesian inference, following the new distribution
\begin{equation}
    \label{eq:HBM1-bayes}
    p(\Theta \mid x) \propto p(x \mid \Theta) p(\Theta).
\end{equation}
This updated distribution takes a simple form. If one observes $N$ datapoints containing $x$ examples of a DO outcome, the updated decision probabilities follow
\begin{equation}
\Theta \mid x \sim \text{Beta}(\alpha + x, \beta + N - x).
\end{equation}
Thus observing many DO outcomes will increase the probability of future DO outcomes. Having demonstrated the logic of Bayesian belief updating in a simple setting, we now turn to our hierarchical model.

\subsubsection{Hierarchical Bayesian Model.} 
As an extension of the Beta-Binomial process, we specify a hierarchical structure for syntactic statistics. Instead of having a single parameter $\Theta$ giving the decision bias, we posit that each verb has an individual decision bias $\phi_v$, which is in turn sampled from a global decision bias variable $\Theta$ that ties the individual verb biases together. Specifically, we posit a generative model
\begin{align}
    \Theta &\sim \mathrm{Beta}(1, 1) \label{HBM2-prior1} \\
    \phi_v &\sim \mathrm{Beta}(\alpha \Theta, \alpha(1-\Theta)) \label{HBM2-prior2}\\
    x_v &\sim \mathrm{Binomial}(\phi_v). \label{HBM2-likelihood}
\end{align}
In this generative model, a global decision bias variable $\Theta$ is sampled from a distribution $\text{Beta}(1,1)$ (which is uniform on the interval $[0,1]$), then for each verb $v$, an individual decision bias $\phi_v$ is sampled from a Beta distribution parameterized by $\Theta$. Here the parameter $\alpha > 0$ represents the relative importance of the global prior $\Theta$ in determining the verb-specific biases $\phi_v$. 

The hierarchical structure of the generative model means that experience with one verb can cause changes not only to that verb's decision bias, but also to the global decision bias, which in turn affects other verbs. Given counts of $x_v$ DO outcomes for a specific verb $v$, when we are considering the update to the statistics for the currently-observed verb $v$, then the update is given by Bayes' rule applied to $\phi_v$ alone:
\begin{align}
p(\phi_v \mid x_v) &\propto p(x_v \mid \phi_v) p(\phi_v).
\end{align}
When we are considering the update to the variable $\phi_w$ for a \emph{different} verb $w \neq v$, then the update is mediated through the updated global statistics $\Theta$:
\begin{align}
\label{HBM2-marginal}
p(\phi_w \mid x_v) &= \int p(\phi_w \mid \Theta)p(\Theta \mid x_v) \text{ d}\Theta.
\end{align}
Therefore, information flows both bottom-up and top-down in the HBM. When the new data $x_v$ for verb $v$ is encountered, it impacts the verb-specific parameter $\phi_v$ and the effect goes bottom up to impact the higher-level $\Theta$. Since this abstract $\Theta$ governs and constrain all the verb-specific parameters, as a top-down process, the effect on $\Theta$ in turn influences other $\phi_w$. In this way, learning from the data of one verb can be generalized to others. We implement the learning of HBM using \texttt{WebPPL} \cite{dippl}.

%

\begin{figure*}[htp]
    \centering
    \includegraphics[width=0.9\linewidth]{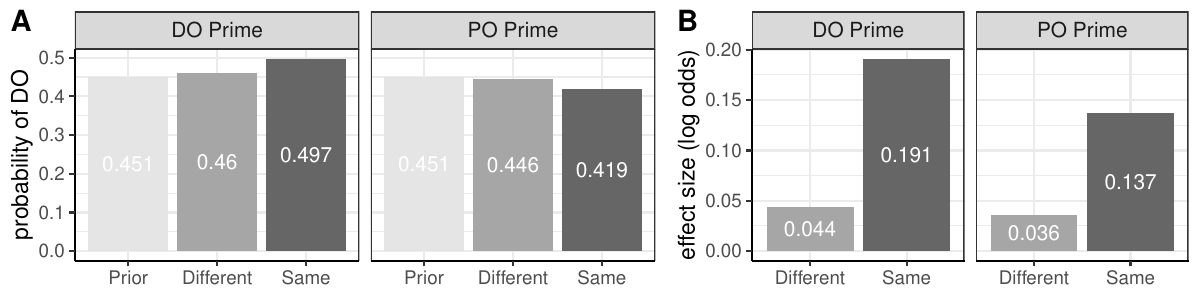}
    \caption{Results of Simulation~1. \textit{Panel A}: Model-estimated average prior and posterior probability of DO for the target verbs in \citeA{pickering1998representation}; \texttt{Same} refers to the condition with verb overlap between the prime and the target; \texttt{Different} refers to the condition without verb overlap. \textit{Panel B}: Model-predicted priming effect size, calculated as the difference of log-odds between the posterior and the prior for DO.}
    \label{fig:result-sim1}
\end{figure*}

\section{Model Evaluation}

In two simulations, we evaluate our model on the experimental materials of \citeA{pickering1998representation}'s Experiment~1, aiming to capture the three empirical properties of syntactic priming outlined earlier. The original experiment is in a trial-to-trial production priming paradigm using English ditransitives. There are 32 items, each consisting of a prime and a target as in (\ref{PB98}) below. At the prime, participants are presented with the partial input of a ditransitive sentence until the first post-verbal object noun phrase, and are asked to complete the sentence. At the target, participants are asked to complete another partial sentence input that includes a subject noun phrase and a ditransitive verb. The prime sentences are manipulated in a 2$\times$2 design. First, the prime is biased towards either DO or PO structure. Second, verb overlap is manipulated such that the prime either shares or does not share the same verb as the target. 

\begin{enumerate}[(1)]
\setcounter{enumi}{\value{list-counter}}
    \item \label{PB98} Sample stimuli from \citeA{pickering1998representation}
    \begin{enumerate}[a.]
        \item \textit{Prime}: \\
            The racing driver showed the torn overall... \\
            The racing driver showed the helpful mechanic... \\
            The racing driver gave the torn overall...\\
            The racing driver gave the helpful mechanic...
        \item \textit{Target}: The patient showed...
    \end{enumerate}
\setcounter{list-counter}{\value{enumi}}
\end{enumerate}

\subsection{Simulation 1: Lexical Boost and Inverse Frequency}

This first simulation aims to capture both the lexical boost and the inverse frequency effect. First, we expect the materials with verb overlap in \citeA{pickering1998representation} to generate a stronger priming effect in the model output, compared to the condition without verb overlap. Second, we expect the model to generate stronger priming effect for the ditransitive structure that is favored \emph{a priori} before priming.

\subsubsection{Estimating the prior.} 
We first construct a dataset to represent the participant's language experience prior to the priming experiment.\footnote{In Bayesian learning, this prior dataset corresponds to pseudocounts. The larger the size of the prior data, the smaller the priming effect for the same amount of exposure data.} As summarized in Table~\ref{tab:prior-data}, this dataset contains 100 datapoints distributed across the nine verbs used in the original experiment. Each data point represents an instance of verb use in one of the two ditransitive structures (e.g. \texttt{<give:DO>}, \texttt{<show:PO>}). The frequency of each verb and their relative use of DO/PO are specified based on counts from the British National Corpus \cite{zhou2023affects, yi2019semantic}.\footnote{The verb frequencies in Table~\ref{tab:prior-data} are sampled from a multinomial distribution with the parameter $N=100$ and the parameter $\mathbf{p}$ corresponding to the verb frequencies in the corpus. The DO/PO frequency for each verb is generated in a similar way but from a binomial distribution.} The model starts with the global bias $\Theta$ and verb-specific biases $\phi_v$ derived from this dataset following the procedures above. This initial state is the prior distribution which will be compared with the updated posterior distribution after exposure to the priming data.

\begin{table}[htp]
    \centering
    \begin{tabular}{lrrr}
    \toprule
       Verb  & DO Freq. & PO Freq. & Verb Freq.\\
       \midrule
       give  & 51 & 20 & 71 \\
       show  & 1  & 3  & 4  \\
       send  & 5  & 8  & 13 \\
       lend  & 1  & 0  & 1  \\
       hand  & 0  & 3  & 3  \\
       loan  & 0  & 0  & 0  \\
       offer & 2  & 4  & 6  \\
       sell  & 0  & 2  & 2  \\
       post  & 0  & 0  & 0  \\
       \textit{in total} & 60 & 40 & 100 \\
    \bottomrule
    \end{tabular}
    \caption{Verb frequencies and counts of DO/PO ditransitives used to form the model's prior distribution.}
    \label{tab:prior-data}
\end{table}

\subsubsection{Inferring the posterior.} 
Starting with the prior defined above, the model infers a posterior based on the exposure data presented in the primes in \citeA{pickering1998representation}. 
The posterior learned by the model is a joint distribution for both the global and verb-specific decision biases. However, the parameters that can be directly measured from the behavioral data are the verb-specific parameters, that is, the proportion of DO/PO use associated with the specific verbs in the target sentence of each item. Therefore, for the target verb in each item, we obtain the verb-specific posterior through marginalization as given in Eq.~\ref{HBM2-marginal}.

\subsubsection{Result.}
Figure~\ref{fig:result-sim1}A shows the average DO probability for the target verbs in each condition. The control condition shows the prior DO probability before priming. Although there are more DOs in the prior data based on the raw frequency in Table~\ref{tab:prior-data}, the model infers a global bias slightly against DO (i.e. $p(\text{DO})<0.5$), because the DO preference in the raw frequency is mainly driven by one single frequent verb \emph{give} in the prior dataset. After updating based on the prime data, we find a priming effect in the form of an increase in DO probability with DO primes and a decrease in DO probability with PO primes. Figure~\ref{fig:result-sim1}B shows the priming effect size more directly by calculating the difference of log-odds between the prior and the posterior of DO for each target verb.

Consistent with lexical boost, the model predicts a stronger priming effect when the prime and the target share the same verb. Consistent with the inverse frequency effect, the model predicts that DO primes---which is the structure less preferred in the prior---should elicit a stronger priming effect.


\subsubsection{Discussion.} 

The model's ability to capture the inverse frequency effect comes with Bayesian learning in general, where events that are more surprising induce larger updates \cite{courville2006bayesian}.
More crucially, the model is also able to capture lexical boost. In fact, in the Bayesian framework, a better way to interpret the effect of verb (non-)overlap is \emph{lexical transfer}, where the learning effect on one verb is generalized to others. It is worth noting that although our model successfully captures the qualitative conclusions in \citeA{pickering1998representation}, it is difficult to directly compare the model-predicted results with the empirical data, because experimental work rarely reports the baseline pre-priming DO probability.

\subsection{Simulation 2: Asymmetrical Decay}

In Simulation~2, we aim to model the asymmetrical decay in syntactic priming, where lexical boost decays more rapidly than the abstract priming effect without verb overlap. The prior estimation follows the procedure identical to Simulation~1, using the prior data in Table~\ref{tab:prior-data}. 

The simulation is conducted in two groups, an exposure and a control group. For the exposure group, the model is first conditioned on the prime sentence. For the control group, there is no conditioning on prime sentences. Then, both groups are exposed to two batches of additional data, each containing 100 samples from the frequency distribution in Table~\ref{tab:prior-data}. These post-priming data reflect the average effect of post-priming trials and experience. The priming effect size is then calculated as the difference of log-odds for DO between the model-inferred posterior in the exposure group and the one in the control group.

\begin{figure}[htp]
    \centering
    \includegraphics[width=0.95\linewidth]{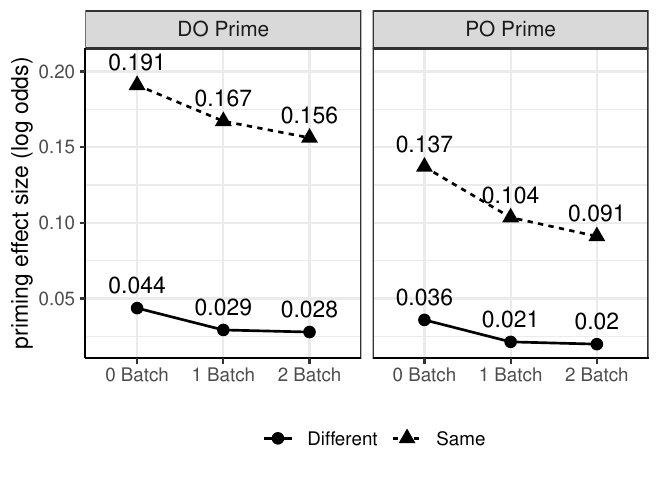}
    \caption{Simulation~2 model-predicted priming effect size as a function of the number of additional batches of post-priming data. Effect size calculated as in Simulation~1.}
    \label{fig:result-sim2}
\end{figure}

\subsubsection{Result.} 
Figure~\ref{fig:result-sim2} shows the model-predicted priming effect size as a function of the number of batches of post-priming data. The result with zero batch of post-priming data is the same as the result in Simulation~1 (Figure~\ref{fig:result-sim1}B), and is repeated here for the convenience of comparison. First, as in Simulation~1, the model reproduces the lexical boost and the inverse frequency effect with additional post-priming data. Second, the result indicates a decay of the priming effect in general for both conditions of verb overlap: The priming effect size decreases as more and more post-priming data are taken by the model. Critically, the model also demonstrates an asymmetry of decay: The priming effect size diminishes more than twice as rapidly in the condition with verb overlap compared to the condition without verb overlap.

\subsubsection{Discussion.} 
As expected, our model successfully captures the asymmetrical longevity of syntactic priming, such that the lexical boost effect decays more rapidly than the priming effect without verb overlap. To our knowledge, the decay of syntactic priming and the distinction between long- and short-term priming has rarely been explained from the perspective of implicit learning, and has actually been considered one of the major limitations of the implicit learning account. However, the result of the current simulation demonstrates that at least some aspects of the long-/short-term distinction in priming can be captured by the implicit learning account in principle. We discuss the potential mechanism of how the HBM can capture the asymmetrical decay of priming below in General Discussion.

\section{General Discussion}

To sum up, in two simulations, we evaluate HBM against the materials from \citeA{pickering1998representation}'s priming experiment in English ditransitives. The model successfully captures lexical boost, the inverse frequency effect, and asymmetrical decay.

\subsection{How abstract is syntactic priming?}

One of the major questions under debate is: How abstract is syntactic priming? Previous studies observe that priming can be somewhat context-independent, for example, without the alignment of semantic features \cite{bock1990framing, bock1992conceptual} or the specific phonetic content of the morphosyntactic markers \cite{bock1989closed, tree1999building}. However, in terms of whether priming is verb-independent, the empirical evidence presents a mixed picture. Although there is some evidence for verb-independent syntactic priming from the production side, studies focused on comprehension tend to find that verb overlap is required for a reliable priming effect \cite{tooley2010syntactic}.

In the current study, by positing a verb-specific level of statistics directly matching the structure of verb-specific input data (i.e. \texttt{<give:DO>}), our model assumes that the major locus of syntactic priming lies in the verb-specific biases. At least for the scenario of ditransitives in the current study, this means that what has been directly learned from the input data is the verb-specific statistics, and that there is a lexical basis of syntactic priming where it is the syntactic knowledge stored in the specific verb items that is directly primed. The abstract higher-level priming, under the framework of HBM, can be viewed as an emergent effect that is driven by the verb-specific effects. This idea of lexical-driven syntactic priming is in fact similar to the assumption held by the activation-based mechanism proposed in \citeA{pickering1998representation}, where the activation of verb lemmas is first boosted by the prime data, and the activation then spreads to the combinatorial nodes that represent syntactic structures.

Given the assumption that syntactic priming is lexical-driven, \textit{lexical boost} can be reinterpreted as \textit{lexical transfer} by reconsidering the role of verb overlap. If priming is assumed to be operated at an abstract level, it would be reasonable to interpret the role of verb overlap as providing additional processing cues to facilitate the selection of an abstract syntactic structure \cite{reitter2011computational}, resulting in a lexical boost. However, if priming is lexical-driven, then the effect of lexical boost, as another side of the same coin, can be interpreted as a weakening effect of verb misalignment. This reinterpretation naturally fits the mechanism of knowledge generalization in HBM: The verb-specific input data not only impacts the verb-specific knowledge, but also goes bottom-up to impact the abstract verb-independent knowledge, which in turn influences other verb-specific knowledge. The magnitude of this generalized priming should be much smaller, not only because the bottom-up effect on the higher-level statistics is counterbalanced by other lower-level statistics, but also because other verb-specific statistics have their own biases.

\subsection{Decay as interference-based unlearning}

What is the mechanism for the decay of priming? In the activation-based account, decay is temporal and in a radioactive fashion. In the implicit learning account, decay is simply underspecified, with no mechanism to unlearn the prime data without any additional input. As intuitive as the time-related decay seems to be, it is empirically difficult to tease apart from another mechanism of decay, which is the interference-based decay. In fact, although the learning literature does not pay much attention to the interference-based decay, studies of working memory has been gradually converged on the claim that memory interference plays a much more important role than temporal decay for many memory mechanisms, such as memory retrieval \cite{lewis2005activation,vasishth2019computational}.

The interference-based mechanism in fact can be a potential way to ``unlearn'' the primed knowledge for implicit learning. In line with the idea that statistical learning spans over life time \cite{chang2006becoming}, the cognitive system keeps self-updating for the incoming data received after the primes. That being said, for participants in a priming experiment, learning is still ongoing during filler items \cite{branigan1999syntactic} or in the naturalistic communication environment between experiment sessions \cite{kaschak2011long}. This post-priming learning can wash out the original priming effect, manifested as a decay over time.

This interference-based unlearning naturally provides a mechanism for asymmetrical decay in HBM. The lower-level statistics, due to the sparsity of data, can be fairly susceptible to a small amount of incoming data. The higher-level statistics, in contrast, is much more stable as it collects information from all the lower-level statistics. Since the verb-independent priming results from the top-down processing governed by the higher-level statistics, its effect is more stable and less susceptible to the interference from the post-priming data. In another word, although the abstract verb-independent knowledge is more difficult to prime, once it has indeed been primed, it is also more difficult to unprime. 

Although we show that HBM can demonstrate the decay of priming in principle, there are still some aspects of decay that can be hardly explained purely from the perspective of implicit learning as a computational theory. First, although HBM captures the qualitative pattern that lexical boost decays faster than verb-independent priming, many studies find that lexical alignment can completely lose its benefit after decay \cite{hartsuiker2008syntactic}. This pattern cannot be reproduced by HBM, where there is always a stronger priming effect with verb overlap. Second, sampling from prior data in the post-priming stage also has its limitation, especially given that the intervening filler items in the experiments of some studies are in a construction differing from the one represented by the statistics in our model. Model architectures that explicitly include a network of constructions as part of the structure of syntactic knowledge, in which (for example) transitive sentences have some relation to ditransitives, may account for the observed decay of syntactic priming when post-priming trials involve sentences in apparently unrelated constructions.

\section{Conclusion}

In the current study, we propose a hierarchical Bayesian model for syntactic priming. The model represents syntactic knowledge as hierarchical syntactic statistics, with two levels of abstraction to represent the verb-specific biases and the abstract verb-independent bias for syntactic decisions. As a learning model, HBM is able to capture the three properties of syntactic priming: the lexical boost, the inverse frequency effect, and the asymmetrical decay. This result expands the explanatory power of the implicit learning account to cover more empirical properties of syntactic priming that tend to be accounted for through residual activation, such as the lexical boost and at least some aspects of the short-term priming. 



\bibliographystyle{apacite}

\setlength{\bibleftmargin}{.125in}
\setlength{\bibindent}{-\bibleftmargin}

\bibliography{weijie_references}

\end{document}